\documentclass[11pt]{article}
\usepackage{acl2013}
\usepackage{times}
\usepackage{latexsym}
\usepackage{amsmath}
\usepackage{multirow}
\usepackage{url}
\usepackage[normalem]{ulem}
\usepackage{color}
\usepackage[usenames,dvipsnames]{xcolor}
\usepackage{enumitem}
\usepackage[lined,boxed,commentsnumbered]{algorithm2e}
\usepackage{graphicx}

\title{A Sentence Compression Based Framework to Query-Focused Multi-Document Summarization}

\author{Lu Wang$^{1}$ ~ Hema Raghavan$^{2}$ ~ Vittorio Castelli$^{2}$ ~ Radu Florian$^{2}$ ~ Claire Cardie$^{1}$\\
$^{1}$Department of Computer Science, Cornell University, Ithaca, NY 14853, USA\\
{\tt \{luwang, cardie\}@cs.cornell.edu}\\
$^{2}$IBM T. J. Watson Research Center, Yorktown Heights, NY 10598, USA\\
{\tt \{hraghav, vittorio, raduf\}@us.ibm.com} \\
}

\begin{document}
\maketitle

\begin{abstract}
\fontsize{10}{12}\selectfont
We consider the problem of using sentence compression techniques to facilitate query-focused multi-document summarization. We present a sentence-compression-based framework for the task, and design a series of learning-based compression models built on parse trees. An innovative beam search decoder is proposed to efficiently find highly probable compressions. Under this framework, we show how to integrate various indicative metrics such as linguistic motivation and query relevance into the compression process by deriving a novel formulation of a compression scoring function. Our best model achieves statistically significant improvement over the state-of-the-art systems on several metrics (e.g.~8.0\% and 5.4\% improvements in ROUGE-2 respectively) for the DUC 2006 and 2007 summarization task.
\end{abstract}

\section{Introduction}
The explosion of the Internet clearly 
warrants the development of techniques for organizing and 
presenting information to users in an effective way. 
Query-focused multi-document summarization (MDS) methods have been 
proposed as one such technique and have attracted
significant attention in recent years. The goal of query-focused 
MDS is to synthesize a brief (often fixed-length) and
well-organized summary from a set of topic-related documents that
answer a complex question or address a topic statement.  
The resulting summaries, in turn, can  support a number of information
analysis applications including open-ended question answering,
recommender systems, and summarization of search engine results.
As further evidence of its importance, the Document Understanding
Conference (DUC) has used query-focused MDS as its main task since
2004 to foster new research on automatic
summarization in the context of users' needs. 

To date, most top-performing systems for multi-document 
summarization---whether query-specific or not---remain largely {\em extractive}: 
their summaries are comprised exclusively
of sentences selected directly from the documents to be summarized~\cite{Erkan:2004:LGL:1622487.1622501,Haghighi:2009:ECM:1620754.1620807,Celikyilmaz:2011:DTC:2002472.2002535}. Despite their simplicity, extractive approaches have some disadvantages. 
First, lengthy sentences that are partly relevant are either excluded 
from the summary or (if selected) can block the selection of 
other important sentences, due to summary length constraints. 
%
In addition, when people write summaries, they tend to abstract the content and 
seldom use entire sentences taken verbatim from the original documents. 
In news articles, for example, most sentences are lengthy and contain 
both potentially useful information for a summary as well as unnecessary 
details that are better omitted. Consider the following DUC query as input 
for a MDS system:\footnote{From DUC 2005, query for topic d422g.} ``{\em In what ways 
have stolen artworks been recovered? How often are suspects arrested or prosecuted 
for the thefts?}'' One manually generated summary includes the following sentence 
but removes the bracketed words in gray:\\\\
\begin{small}
A man suspected of stealing a million-dollar collection of [\textcolor{Gray}{hundreds of ancient}] Nepalese and Tibetan art objects in New York [\textcolor{Gray}{11 years ago}] was arrested [\textcolor{Gray}{Thursday at his South Los Angeles home, where he had been hiding the antiquities, police said}].\\
\end{small}

\noindent
In this example, the compressed sentence is relatively more succinct and readable 
than the original (e.g.~in terms of  Flesch-Kincaid Reading Ease Score \cite{citeulike:7802029}). Likewise, removing information irrelevant to the query (e.g.~``11 years ago", ``police said") 
is crucial for query-focused MDS.

Sentence compression techniques~\cite{Knight:2000:SSS:647288.721086,Clarke:2008:GIS:1622655.1622667} are the standard for producing a compact and grammatical version of a sentence while preserving relevance, and prior research (e.g.~\newcite{Lin:2003:ISP:1118935.1118936}) has demonstrated their potential usefulness for generic document summarization. Similarly, strides have been made to incorporate sentence compression into query-focused MDS systems \cite{Zajic2006}. 
Most attempts, however,  
fail to produce better results than those of the best systems built on pure 
extraction-based approaches that use no sentence compression.

In this paper we investigate the role of sentence compression
techniques for query-focused MDS.
We extend existing work in the area first by investigating
the role of {\em learning-based} sentence compression techniques.
In addition, we design three types of approaches to sentence-compression---{\it rule-based}, {\it sequence-based} and
{\it tree-based}---and examine them within our
compression-based framework for query-specific MDS.
Our
top-performing sentence compression algorithm incorporates measures of query
relevance, content importance, redundancy and language quality, among
others. 
Our tree-based methods rely on a scoring function that allows for easy and flexible tailoring of
sentence compression to the summarization task, ultimately resulting in significant improvements for MDS, while at the same time remaining competitive with existing methods in terms of sentence compression, as discussed next.

We evaluate the summarization models on the standard Document Understanding
Conference (DUC) 2006 and 2007 corpora \footnote{We believe that we can easily adapt our system for  tasks (e.g.~TAC-08's opinion summarization or TAC-09's update summarization) or  domains (e.g.~web pages or wikipedia pages). We reserve that for future work.}
for query-focused MDS and find that
all of our compression-based summarization models achieve
statistically significantly better performance than the best DUC 2006
systems.
Our best-performing system yields an 11.02 ROUGE-2
score~\cite{Lin:2003:AES:1073445.1073465}, a 8.0\% improvement over the best reported score (10.2~\cite{DavisCS12ICDM}) on the DUC 2006 dataset, and an 13.49 ROUGE-2, a 5.4\% improvement over the best score in DUC 2007 (12.8~\cite{DavisCS12ICDM}).
We also observe substantial improvements over previous systems
w.r.t.~the manual Pyramid~\cite{nenkova-passonneau:2004:HLTNAACL} evaluation
measure (26.4 vs. 22.9~\cite{iiit06});
human annotators furthermore rate our system-generated summaries as
having less redundancy and comparable quality w.r.t.~other linguistic
quality metrics.
With these results we believe we are the first to successfully show that
sentence compression can provide statistically significant
improvements over pure extraction-based approaches for query-focused
MDS.

\section{Related Work}
Existing research on query-focused multi-document summarization (MDS) largely relies on extractive approaches, where systems usually take as input a set of documents and select the top relevant sentences for inclusion in the final summary. A wide range of methods have been employed for this task. For unsupervised methods, sentence importance can be estimated by calculating topic signature words~\cite{Lin:2000:AAT:990820.990892,classy}, combining query similarity and document centrality within a graph-based model~\cite{Otterbacher:2005:URW:1220575.1220690}, or using a Bayesian model with sophisticated inference~\cite{Daume:2006:BQS:1220175.1220214}. \newcite{DavisCS12ICDM} first learn the term weights by Latent Semantic Analysis, and then greedily select sentences that cover the maximum combined weights. Supervised approaches have mainly focused on applying discriminative learning for ranking sentences~\cite{Fuentes:2007:SVM:1557769.1557788}. \newcite{Lin:2011:CSF:2002472.2002537} use a class of carefully designed submodular functions to reward the diversity of the summaries and select sentences greedily. 

Our work is more related to the less studied area of sentence compression as applied to (single) document summarization. \newcite{Zajic2006} tackle the query-focused MDS problem using a “compress-first” strategy: they develop heuristics to generate multiple alternative compressions of all sentences in the original document; these then become the candidates for extraction. This approach, however, does not outperform some extraction-based approaches. A similar idea has been studied for MDS~\cite{Lin:2003:ISP:1118935.1118936,Gillick:2009:SGM:1611638.1611640}, but limited improvement is observed over extractive baselines with simple compression rules. Finally, although learning-based compression methods are promising~\cite{Martins:2009:SJM:1611638.1611639,Berg-Kirkpatrick:2011:JLE:2002472.2002534}, it is unclear how well they handle issues of redundancy.

Our research is also inspired by probabilistic sentence-compression approaches, such as the noisy-channel model~\cite{Knight:2000:SSS:647288.721086,Turner:2005:SUL:1219840.1219876}, and its extension via synchronous context-free grammars (SCFG)~\cite{DBLP:journals/jcss/AhoU69,Lewis:1968:ST:321466.321477} for robust probability estimation~\cite{galley-mckeown:2007:main}. Rather than attempt to derive a new parse tree like \newcite{Knight:2000:SSS:647288.721086} and \newcite{galley-mckeown:2007:main}, we learn to safely remove a set
of constituents in our parse tree-based compression model while preserving grammatical structure and essential content. Sentence-level compression has
also been examined via a discriminative model \newcite{McDonald06}, and \newcite{Clarke:2008:GIS:1622655.1622667} also incorporate discourse information by using integer linear programming.

\section{The Framework}
We now present our query-focused MDS framework consisting of three steps: Sentence Ranking, Sentence Compression and Post-processing. 
First, sentence ranking determines
the importance of each sentence given the query. 
Then, a sentence compressor 
iteratively generates the most likely succinct versions of the ranked sentences, 
which are cumulatively added to the summary, until 
a length limit is reached.
Finally, the post-processing stage applies coreference resolution and sentence reordering to build the summary.

\paragraph{Sentence Ranking.}
\label{sentRank}
\begin{table}
    {\scriptsize
    \setlength{\baselineskip}{0pt}
    \begin{tabular}{|l|}
    \hline
    {\bf Basic Features}\\ \hline
    relative/absolute position\\
    is among the first 1/3/5 sentences?\\
    number of words (with/without stopwords)\\
    number of words more than 5/10 (with/without stopwords)\\
    \hline
    
    {\bf Query-Relevant Features}\\ \hline
    unigram/bigram/skip bigram (at most four words apart) overlap\\
    unigram/bigram TF/TF-IDF similarity\\
	mention overlap\\
	subject/object/indirect object overlap\\
	semantic role overlap\\
	relation overlap\\
    \hline
    
    {\bf Query-Independent Features}\\ \hline
	average/total unigram/bigram IDF/TF-IDF\\
	unigram/bigram TF/TF-IDF similarity with the centroid of the cluster\\
	average/sum of sumBasic/SumFocus~\cite{sumbasic}\\
	average/sum of mutual information\\
	average/sum of number of topic signature words~\cite{Lin:2000:AAT:990820.990892}\\
	basic/improved sentence scorers from~\newcite{classy}\\
    \hline
    
    {\bf Content Features}\\ \hline    
	contains verb/web link/phone number?\\
	contains/portion of words between parentheses\\
    \hline    
    
    \end{tabular}
    }
    \vspace{-1mm}
    \caption{\fontsize{10}{12}\selectfont Sentence-level features for sentence ranking.}
    \label{tab:featureSentRanking}

\end{table}

This stage aims to rank sentences in order of relevance to the query. Unsurprisingly, ranking algorithms have been successfully applied to this task. We experimented with two of them -- Support Vector Regression (SVR)~\cite{DBLP:conf/nips/1996N} and LambdaMART~\cite{NIPS2006_574}. The former has been used previously for MDS~\cite{Ouyang:2011:ARM:1945081.1945172}. LambdaMart on the other hand has shown considerable success in information retrieval tasks~\cite{burges_2010}; we are the first to apply it to summarization. 
For training, we use 40 topics (i.e.~queries) from the DUC 2005 corpus~\cite{duc2005} along with their manually generated abstracts.  As in previous work~\cite{Shen11,Ouyang:2011:ARM:1945081.1945172}, we  
use the ROUGE-2 score, which measures bigram overlap between a sentence and the abstracts, as the objective for regression.

%
While space limitations preclude a longer discussion of the full feature set (ref. Table~\ref{tab:featureSentRanking}), we describe next the 
query-relevant features used for sentence ranking as these are the most important for our
summarization setting.
The goal of this feature subset is to determine the similarity between the query and each candidate sentence. 
When computing similarity, we remove stopwords as well as the words ``discuss, describe, specify, explain, identify, include, involve, note" that are adopted and extended from~\newcite{classy}. Then we conduct simple query expansion based on the title of the topic and cross-document coreference resolution. Specifically, we first add the words from the topic title to the query. And for each mention in the query, we add other mentions within the set of documents that corefer with this mention. Finally, we compute two versions of the features---one based on the original query and another on the expanded one. 
We also derive the semantic role overlap and relation instance overlap between the query and each sentence. Cross-document coreference resolution, semantic role labeling and relation extraction are accomplished via the methods described in Section~\ref{expSetup}.

\paragraph{Sentence Compression.}
As the main focus of this paper, we propose three types of compression methods, described in detail in Section~\ref{sentComp} below.

\paragraph{Post-processing.}
Post-processing performs \emph{coreference} \emph{resolution} and \emph{sentence} \emph{ordering}. We replace each pronoun with its referent unless they appear in the same sentence. 
For sentence ordering, each compressed sentence is assigned to the most similar (tf-idf) query sentence. Then a Chronological Ordering algorithm~\cite{Barzilay:2002:ISS:1622810.1622812} sorts the sentences for each query based first on the time stamp, and then the position in the source document.



\section{Sentence Compression}
\label{sentComp}
Sentence compression is typically formulated as the problem of removing secondary information from a sentence while maintaining its grammaticality and semantic structure~\cite{Knight:2000:SSS:647288.721086,McDonald06,galley-mckeown:2007:main,Clarke:2008:GIS:1622655.1622667}. We leave other rewrite operations, such as paraphrasing and reordering, for future work.
Below we describe the sentence compression approaches developed in this research: \textsc{Rule-based Compression}, \textsc{Sequence-based Compression}, and \textsc{Tree-based Compression}.

\subsection{Rule-based Compression}
\begin{table*}
\centering
    {\fontsize{8.3}{10}\selectfont
    \setlength{\baselineskip}{0pt}
    \begin{tabular}{|l|l|}
    \hline
    \textbf{Rule}&\textbf{Example}\\ \hline
    Header & [\textcolor{Gray}{MOSCOW , October 19 ( Xinhua ) --}] Russian federal troops Tuesday continued...\\ \hline
    Relative dates & ...Centers for Disease Control confirmed [\textcolor{Gray}{Tuesday}] that there was...\\ \hline
    Intra-sentential attribution & ...fueling the La Nina weather phenomenon, [\textcolor{Gray}{the U.N. weather agency said}].\\ \hline
    Lead adverbials & [\textcolor{Gray}{Interestingly}], while the Democrats tend to talk about...\\ \hline
    Noun appositives & Wayne County Prosecutor [\textcolor{Gray}{John O'Hara}] wanted to send a message...\\ \hline
    Nonrestrictive relative clause & Putin, [\textcolor{Gray}{who was born on October 7, 1952 in Leningrad}], was elected in the presidential election...\\ \hline
    Adverbial clausal modifiers &  [\textcolor{Gray}{Starting in 1998}], California will require 2 per cent of a manufacturer...\\
    (Lead sentence) &  [\textcolor{Gray}{Given the short time}], car makers see electric vehicles as...\\ \hline
    Within Parentheses &  ...to Christian home schoolers in the early 1990s [\textcolor{Gray}{(www.homecomputermarket.com)}].\\
    \hline
    \end{tabular}
    }
    \vspace{-3mm}
    \caption{\fontsize{10}{12}\selectfont Linguistically-motivated rules for sentence compression. The grayed-out words in brackets
    are removed.}
    \label{tab:lingRules}
\end{table*}

\newcite{Turner:2005:SUL:1219840.1219876} have shown that applying hand-crafted rules for trimming sentences can improve both content and linguistic quality. Our rule-based approach extends existing work~\cite{classy,sumbasic} to create the linguistically-motivated compression rules of Table~\ref{tab:lingRules}.
To avoid ill-formed output, we disallow  compressions of more than 10 words by each rule.

\subsection{Sequence-based Compression}

\begin{table}
    {\scriptsize
    \setlength{\baselineskip}{0pt}
    
    \hspace{2mm}
    \begin{tabular}{|l|l|}
    \hline
    
    \underline{\bf Basic Features} & \underline{\bf Syntactic Tree Features}\\ 
    first 1/3/5 tokens (toks)? & POS tag\\
    last 1/3/5 toks? & parent/grandparent label\\
    first letter/all letters capitalized? & leftmost child of parent?\\
    is negation? & second leftmost child of parent?\\
    is stopword? & is headword?\\
    
    \underline{\bf Dependency Tree Features} & in NP/VP/ADVP/ADJP chunk?\\ 
    dependency relation (dep rel) & \underline{\bf Semantic Features}\\
    parent/grandparent dep rel & is a predicate?\\
    is the root? & semantic role label\\
    has a depth larger than 3/5? & \\
    \hline
    
%
%
%
%
    \end{tabular}
    
    \hspace{2mm}
	\begin{tabular}{|p{67.5mm}|}
	\hline
    \underline{\bf Rule-Based Features}\\
	For each rule in Table~\ref{tab:lingRules} , we construct a corresponding feature to indicate whether the token is identified by the rule.\\
	\hline
    \end{tabular}
    
    }
    \vspace{-1mm}
    \caption{\fontsize{10}{12}\selectfont Token-level features for sequence-based compression.}
    \label{tab:featureSeqComp}

\end{table}

As in \newcite{McDonald06} and~\newcite{Clarke:2008:GIS:1622655.1622667}, our sequence-based compression model makes a binary ``keep-or-delete" decision for each word in the sentence. 
In contrast, however, we view compression as a sequential tagging problem and make use of linear-chain Conditional Random Fields (CRFs)~\cite{Lafferty:2001:CRF:645530.655813} to select the most likely compression. We represent each sentence as a sequence of tokens, $X=x_{0}x_{1}\ldots x_{n}$, and   generate a sequence of labels, $Y=y_{0}y_{1}\ldots y_{n}$, that encode which tokens are kept, using a BIO label format: \{\textsc{B-Retain} denotes the beginning of a retained sequence, \textsc{I-Retain} indicates tokens ``inside" the retained sequence,  \textsc{O} marks tokens to be removed\}.

%
%

The CRF model is built using the features shown in Table~\ref{tab:featureSeqComp}. ``Dependency Tree Features"  encode the grammatical relations in which each word is involved as a dependent. For the ``Syntactic Tree", ``Dependency Tree" and ``Rule-Based" features, we also include features for the two words that precede and the two that follow the current word. Detailed descriptions of the training data and experimental setup are in Section~\ref{expSetup}.

During inference, we find the maximally likely sequence $Y$ according to a CRF with parameter $\theta$ ($Y=\arg\max_{Y'} P(Y'|X;\theta)$), while simultaneously enforcing the rules of Table~\ref{tab:lingRules} to reduce the hypothesis space and encourage grammatical compression. To do this, we encode these rules as features for each token, and whenever these feature functions fire, we restrict the possible label for that token to ``O".
 
\subsection{Tree-based Compression}
\begin{figure}
\hspace{5mm}
\includegraphics[width=65mm,height=35mm]{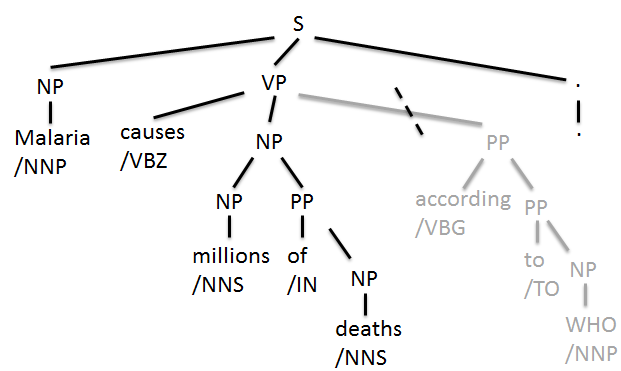}
\vspace{-2mm}
\caption{\fontsize{10}{12}\selectfont Diagram of tree-based compression. The nodes to be dropped are grayed out. In this example, the root of the gray subtree (a ``PP") would be labeled \textsc{Remove}. Its siblings and parent are labeled \textsc{Retain} and \textsc{Partial}, respectively. The trimmed tree is realized as ``{\it Malaria causes millions of deaths.}"}
\label{fig:parsetree}
\end{figure}

Our tree-based compression methods are in line with syntax-driven approaches~\cite{galley-mckeown:2007:main}, where operations are carried out on parse tree constituents. Unlike previous work~\cite{Knight:2000:SSS:647288.721086,galley-mckeown:2007:main}, we do not produce a new parse tree, but focus on learning
to identify the proper set of constituents to be removed. 
In particular, when a node is dropped from the tree, all words it subsumes will be deleted
from the sentence. 

Formally, given a parse tree $T$ of the sentence to be compressed and a tree traversal algorithm, $T$ can be presented as a list of ordered constituent nodes, $T=t_{0}t_{1}\ldots t_{m}$. Our objective is to find a set of labels, $L=l_{0}l_{1}\ldots l_{m}$, where $l_{i}$ $\in$ \{\textsc{Retain}, \textsc{Remove}, \textsc{Partial}\}. 
\textsc{Retain} (\textsc{Ret}) and \textsc{Remove} (\textsc{Rem}) denote whether the node $t_{i}$ is retained or removed. \textsc{Partial} (\textsc{Par}) means $t_{i}$ is partly removed, i.e.~at least one child subtree of $t_{i}$ is dropped.

Labels are identified, in order, according to the tree traversal algorithm. Every node
label needs to be {\em compatible} with the labeling history: 
given a node $t_{i}$, and a set of labels $l_{0}\ldots l_{i-1}$ predicted for nodes $t_{0}\ldots t_{i-1}$, $l_{i}=$\textsc{Ret} or $l_{i}=$\textsc{Rem} is {\em compatible} with the history when
all children of $t_{i}$ are labeled as \textsc{Ret} or \textsc{Rem}, respectively; 
$l_{i}=$\textsc{Par} is {\em compatible} when $t_{i}$ has at least two descendents $t_{j}$ and $t_{k}$ 
($j<i$ and $k<i$), 
one of which is \textsc{Ret}ained and the other, \textsc{Rem}oved.
As such, the 
root of the gray subtree in Figure~\ref{fig:parsetree} is 
labeled as \textsc{Rem}; its left siblings as \textsc{Ret}; its parent as \textsc{Par}.

As the space of possible compressions is exponential in the number of leaves in the parse tree, instead of looking for the globally optimal solution, we  use beam search  to find a set of highly likely 
compressions and employ a language model trained on a large corpus for evaluation.
%

\begin{figure}
\hspace{-2mm}
\includegraphics[width=80mm,height=35mm]{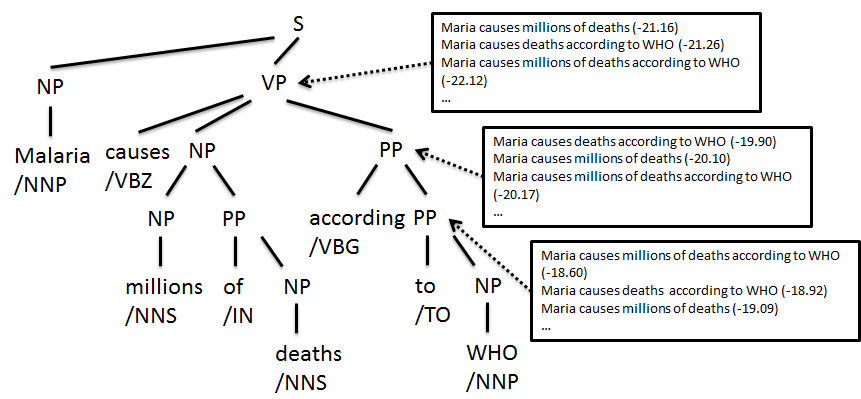}
\vspace{-7mm}
\caption{\fontsize{10}{12}\selectfont Example of beam search decoding. For postorder traversal, the three nodes 
are visited in a bottom-up order. The associated compression hypotheses (boxed) are ranked based on the scores in parentheses. Beam scores for other nodes are omitted.}
\label{fig:beamsearch}
\end{figure}

\paragraph{A Beam Search Decoder.}
The beam search decoder (see Algorithm~\ref{alg:beamSearch}) takes as input the sentence's parse tree $T=t_{0}t_{1}\ldots t_{m}$, an ordering $O$ for traversing $T$ (e.g.~postorder) as a sequence of nodes in $T$, the set $L$ of possible node labels, a scoring function $S$ for evaluating each sentence compression hypothesis, and a beam size $N$. Specifically, $O$ 
is a permutation on the set $\{0, 1, \ldots, m\}$---each element an index onto $T$.
Following $O$, $T$ is re-ordered as $t_{O_{0}}t_{O_{1}}\ldots t_{O_{m}}$, and the decoder considers 
each ordered constituent $t_{O_{i}}$ in turn.
%
In iteration $i$, all existing sentence compression hypotheses are expanded by one node, $t_{O_{i}}$,
labeling it with {\bf all} {\em compatible} labels.
The new hypotheses (usually sub-sentences) are ranked by the scorer $S$ and the top $N$ are preserved to be extended in the next iteration. 
See Figure~\ref{fig:beamsearch} for an example.
\begin{algorithm}
\small
\SetKwInOut{Input}{Input}
\SetKwInOut{Output}{Output}
\Input{parse tree $T$, ordering $O=O_{0}O_{1}\ldots O_{m}$, $L=$\{\textsc{Ret}, \textsc{Rem}, \textsc{Par}\}, hypothesis scorer $S$, beam size $N$}
\Output{$N$ best compressions}
\BlankLine

stack $\leftarrow \Phi$ (empty set)\;
\ForEach{node $t_{O_{i}}$ in $T=t_{O_{0}}\ldots t_{O_{m}}$}{
    \eIf{$i==0$ (first node visited)}{
        \ForEach{label $l_{O_{0}}$ in $L$}{
    		newHypothesis $h'\leftarrow[l_{O_{0}}]$\;
			put $h'$ into Stack\;
		}
	}
	{	
		newStack $\leftarrow \Phi$ (empty set)\;
		\ForEach{hypothesis $h$ in {\normalfont stack}}{
			\ForEach{label $l_{O_{i}}$ in $L$}{
				\If{$l_{O_{i}}$ is compatible}{
					newHypothesis $h'\leftarrow h+[l_{O_{i}}]$\;
					put $h'$ into newStack\;
										
				}
				
			}
		
		}
		stack $\leftarrow$ newStack\;
	}
	Apply $S$ to sort hypotheses in stack in descending order\;
	Keep the $N$ best hypotheses in stack\;

}

\caption{\fontsize{10}{12}\selectfont Beam search decoder.}
\label{alg:beamSearch}
\end{algorithm}

Our {\bf \textsc{Basic} Tree-based Compression} instantiates the beam search decoder with postorder traversal and a hypothesis scorer 
%
that takes a possible sentence compression---a sequence of nodes (e.g.~$t_{O_{0}}\ldots t_{O_{k}}$) and
their labels (e.g.~$l_{O_{0}}\ldots l_{O_{k}}$)---and returns $\sum_{j=1}^{k} \log P(l_{O_{j}}|t_{O_{j}})$
(denoted later as $Score_{Basic}$). 
The probability is estimated by a Maximum Entropy classifier~\cite{Berger:1996:MEA:234285.234289} trained at the constituent level using the features in Table~\ref{tab:featureTreeComp}.  We also apply the rules of Table~\ref{tab:lingRules} during the decoding process. Concretely, if the words subsumed by a node are identified by any rule, we only consider \textsc{Rem} as the node's label.

Given the $N$-best compressions from the decoder, we evaluate the yield of the trimmed trees using a language model 
trained on the Gigaword~\cite{gigaword} corpus and return the compression with the highest probability.
Thus, the decoder is quite flexible --- its learned scoring function allows us to 
incorporate features salient for sentence compression 
while its language model guarantees the linguistic quality of the compressed string.  
In the sections below we consider additional improvements.

\begin{table}
    {\scriptsize
    \setlength{\baselineskip}{0pt}
    \begin{tabular}{|l|l|}
    \hline
    
    \underline{\bf Basic Features} & \underline{\bf Syntactic Tree Features}\\
    projection falls w/in first 1/3/5 toks?$\ast$ & constituent label\\
    projection falls w/in last 1/3/5 toks?$\ast$ & parent left/right sibling label\\
    subsumes first 1/3/5 toks?$\ast$ & grandparent left/right sibling label\\
    subsumes last 1/3/5 toks?$\ast$ & is leftmost child of parent?\\
    number of words larger than 5/10?$\ast$ & is second leftmost child of parent?\\
    is leaf node?$\ast$ & is head node of parent?\\
    is root of parsing tree?$\ast$ & label of its head node\\
    has word with first letter capitalized? & has a depth greater than 3/5/10?\\
	has word with all letters capitalized? & \underline{\bf Dependency Tree Features} \\
	has negation? & dep rel of head node$\dagger$\\
	has stopwords? & dep rel of parent's head node$\dagger$\\
	    
    \underline{\bf Semantic Features} & dep rel of grandparent's head node$\dagger$\\
    the head node has predicate? & contain root of dep tree?$\dagger$\\
   	semantic roles of head node & has a depth larger than 3/5?$\dagger$\\
    
    \hline
%
%
%
%
    \end{tabular}

	\begin{tabular}{|p{76.6mm}|}
	\hline
    \underline{\bf Rule-Based Features}\\
	For each rule in Table~\ref{tab:lingRules} , we construct a corresponding feature to indicate whether the token is identified by the rule.\\
	\hline
    \end{tabular}    
    
    }
    \vspace{-1mm}
    \caption{\fontsize{10}{12}\selectfont Constituent-level features for tree-based compression. $\ast$ or $\dagger$ denote features that are concatenated with every Syntactic Tree feature to compose a new one.}
    \label{tab:featureTreeComp}
\end{table}


\subsubsection{Improving Beam Search}

{\bf \textsc{Context}-aware search} is based on the intuition that predictions on preceding context can be leveraged to facilitate the prediction of the current node. For example, parent
nodes with children that have all been removed (retained) should have a label of 
\textsc{Rem} (\textsc{Ret}). 
%
In light of this, we encode these contextual predictions as additional features of $S$, that is, \textsc{all-children-removed/retained}, \textsc{any-left\-Sib\-ling-re\-moved/re\-tained/partly\_removed}, \textsc{label-of-left-sibling/head-node}.

{\bf \textsc{Head}-driven search} modifies the \textsc{Basic} postorder tree traversal by visiting the head node first at each level, leaving other orders unchanged.
In a nutshell, if the head node is dropped, then its modifiers need not be preserved. We adopt the same features as \textsc{context}-aware search, but remove those involving left siblings. We also add one more feature: \textsc{label-of-the-head-node-it-modifies}.

\subsubsection{Task-Specific Sentence Compression}
\label{scoringTech}


The current scorer $Score_{Basic}$ is
still fairly naive in that it focuses only on features
of the sentence to be compressed. {\em However extra-sentential knowledge
  can also be important 
  for query-focused MDS}. For example, 
  information regarding relevance to the query might lead the decoder to produce
compressions better suited for the summary. Towards this
goal, we construct a compression scoring function---the {\em multi-scorer} (\textsc{Multi})---that allows the
incorporation of multiple task-specific scorers.  Given a
hypothesis at any stage of decoding, which yields a sequence of words
$W=w_{0}w_{1}...w_{j}$, we propose the following component scorers.

\paragraph{Query Relevance.} Query information ought to guide the compressor to identify the relevant content. The query $Q$ is expanded as described in Section~\ref{sentRank}. Let $|W\cap Q|$ denote the number of unique overlapping words between $W$ and $Q$, then $score_{q} = |W\cap Q|/|W|$.

\paragraph{Importance.} A query-independent importance score is defined as the average 
SumBasic~\cite{sumbasic} value in $W$, i.e.~$score_{im} = \sum_{i=1}^{j}SumBasic(w_{i})/|W|$.

\paragraph{Language Model.} We let $score_{lm}$ be the probability of $W$ computed by a language model.

\paragraph{Cross-Sentence Redundancy.} To encourage diversified content, we define a redundancy score to discount replicated content: $score_{red}=1-|W\cap C|/|W|$, where $C$ is the words already selected for the summary.

The {\em multi-scorer} is defined as a linear combination of the component scorers:
Let $\vec{\alpha}=(\alpha_{0}, \ldots, \alpha_{4})$, $0\leq\alpha_{i}\leq1$, $\overrightarrow{score}=(score_{Basic}, score_{q}, score_{im}, score_{lm}, score_{red})$,

\vspace{-2mm}
\begin{equation}
S = score_{multi} = \vec{\alpha  } \cdot \overrightarrow{score}
\end{equation}

\noindent
The parameters $\vec{\alpha}$ are tuned on a held-out tuning set by grid search. We linearly normalize the score of each metric, where the minimum and maximum values are estimated from the tuning data.

\section{Experimental Setup}
\label{expSetup}
We evaluate our methods on the DUC 2005, 2006 and 2007 datasets~\cite{duc2005,duc,duc2007}, each of which is a collection of newswire articles.
50 complex queries (topics) are provided for DUC 2005 and 2006, 35 are collected for DUC 2007 main task. Relevant documents for each query are provided along with 4 to 9 human MDS abstracts.
The task is to generate a summary within 250 words to address the query.
We split DUC 2005 into two parts: 40 topics to train the sentence ranking models, and 10
for ranking algorithm selection and parameter tuning for the multi-scorer. 
DUC 2006 and DUC 2007 are reserved as held out test sets.


{\em Sentence Compression.}
The dataset from~\newcite{Clarke:2008:GIS:1622655.1622667} is used to train the CRF and MaxEnt
classifiers (Section~\ref{sentComp}). It includes 82 newswire articles with one manually produced compression aligned to each sentence. 

{\em Preprocessing.}
Documents are processed by a full NLP pipeline, including token and sentence segmentation, parsing, semantic role labeling, and an information extraction pipeline consisting of mention detection, NP coreference, cross-document resolution, and relation detection~\cite{DBLP:conf/naacl/FlorianHIJKLNR04,DBLP:conf/acl/LuoIJKR04,DBLP:conf/naacl/LuoZ05}.

{\em Learning for Sentence Ranking and Compression.}
We use Weka~\cite{weka} to train a support vector regressor and experiment with various rankers in RankLib~\cite{ranklib}\footnote{Default parameters are used. If an algorithm needs a validation set, we use 10 out of 40 topics.}. As LambdaMART has an edge over other rankers on the held-out dataset, we selected it to produce ranked sentences for further processing.
For sequence-based compression using CRFs, we employ Mallet~\cite{mallet} and integrate the 
Table~\ref{tab:lingRules} rules during inference. NLTK~\cite{nltk} MaxEnt classifiers are used 
for tree-based compression. Beam size is fixed at 2000.\footnote{We looked at various beam sizes
on the heldout data, and observed that the performance peaks around this value.} Sentence
compressions are evaluated by a 5-gram language model trained on Gigaword~\cite{gigaword} 
by SRILM~\cite{srilm}.

%
%

\section{Results}

\begin{table*}[ht]
\centering
    {\small
    \setlength{\baselineskip}{0pt}
    \begin{tabular}{|l|c|c|c|c|c|c|}
    \hline
    &\multicolumn{3}{|c|}{DUC 2006}&\multicolumn{3}{|c|}{DUC 2007}\\\hline
    \textbf{System}&\textbf{C Rate}&\textbf{R-2}&\textbf{R-SU4}&\textbf{C Rate}&\textbf{R-2}&\textbf{R-SU4}\\\hline
    Best DUC system& --  & 9.56 & 15.53 & --  & 12.62 & 17.90\\
    \newcite{DavisCS12ICDM} & --  & 10.2 & 15.2 & -- & 12.8& 17.5\\
    SVR & 100\%  & 7.78 & 13.02 & 100\% & 9.53 & 14.69\\
    LambdaMART & 100\%  & 9.84 & 14.63 & 100\%  & 12.34& 15.62\\
    \hline

    Rule-based & 78.99\%  & 10.62 $\ast\dagger$ & 15.73 $\dagger$ & 78.11\%  & 13.18$\dagger$ & 18.15$\dagger$\\
    Sequence & 76.34\%  & 10.49 $\dagger$ & 15.60 $\dagger$& 77.20\%  & 13.25$\dagger$& 18.23$\dagger$\\
    Tree ({\sc Basic + $Score_{Basic}$}) & 70.48\%  & 10.49 $\dagger$ & 15.86 $\dagger$ & 69.27\%  & 13.00$\dagger$& 18.29$\dagger$\\
    Tree ({\sc Context + $Score_{Basic}$}) & 65.21\%  & 10.55 $\ast\dagger$ & 16.10 $\dagger$ & 63.44\%  & 12.75& 18.07$\dagger$\\
    Tree ({\sc Head + $Score_{Basic}$}) & 66.70\%  & 10.66 $\ast\dagger$ & 16.18 $\dagger$  & 65.05\%  & 12.93& 18.15$\dagger$\\
    Tree ({\sc Head + Multi}) & 70.20\%  & {\bf 11.02} $\ast\dagger$ & {\bf 16.25} $\dagger$ & 73.40\%  & {\bf 13.49}$\dagger$& {\bf 18.46}$\dagger$\\

    \hline
    \end{tabular}
    }

    \caption{\fontsize{10}{12}\selectfont Query-focused MDS performance comparison: C Rate or
    \emph{compression rate} is the proportion of words preserved. R-2 (ROUGE-2) and R-SU4 (ROUGE-SU4) scores are multiplied by 100. ``--" indicates that data is unavailable. 
    {\sc Basic}, {\sc Context} and {\sc Head} represent the basic beam search decoder,
	context-aware and head-driven search extensions respectively. $Score_{Basic}$ and {\sc Multi}
    refer to the type of scorer used. Statistically significant improvements
    ($p<0.01$) over the best system in DUC 06 and 07 are marked with $\ast$. $\dagger$ indicates
    statistical significance ($p<0.01$) over extractive approaches (SVR or LambdaMART). 
    {\sc Head + Multi} outperforms all the other extract- and compression-based systems in R-2.}
    \label{tab:resultROUGE}

\end{table*}

The results in Table~\ref{tab:resultROUGE} use the official ROUGE software with standard 
options\footnote{ROUGE-1.5.5.pl -n 4 -w 1.2 -m  -2 4 -u -c 95 -r 1000 -f A -p 0.5 -t 0 -a -d}  
and report ROUGE-2 (R-2) (measures bigram overlap) and ROUGE-SU4 (R-SU4) (measures unigram
and skip-bigram separated by up to four words).
%
%
We compare our sentence-compression-based methods to the best performing systems based on ROUGE in DUC 2006 and 2007~\cite{iiit06,iiit}, system by \newcite{DavisCS12ICDM} that report the best R-2 score on DUC 2006 and 2007 thus far, and to the purely extractive methods of SVR and LambdaMART. 

Our sentence-compression-based systems (marked with $\dagger$) show statistically significant improvements over pure extractive summarization for both R-2 and R-SU4 (paired {\it t}-test, $p<0.01$). This means our systems can effectively remove redundancy within the summary through compression. Furthermore, our {\sc Head}-driven beam search method with {\sc Multi}-scorer beats all systems on DUC 2006\footnote{The system output from~\newcite{DavisCS12ICDM} is not available, so significance tests are not conducted on it.} and all systems on DUC 2007 except the best system in terms of R-2 ($p<0.01$). Its R-SU4 score is also significantly ($p<0.01$) better than extractive methods, rule-based and sequence-based compression methods on both DUC 2006 and 2007.
Moreover, our systems with learning-based compression have considerable compression rates, indicating their capability to remove superfluous words as well as improve summary quality.

\paragraph{Human Evaluation.}
%
%
%

\begin{table*}[ht]
\centering
    {\small
    \setlength{\baselineskip}{0pt}
    \begin{tabular}{|l|c|c|c|c|c|c|}
    \hline
    \textbf{System} & \textbf{Pyr} & \textbf{Gra} & \textbf{Non-Red} & \textbf{Ref} & \textbf{Foc} & \textbf{Coh}\\ \hline
    Best DUC system (ROUGE) & 22.9$\pm$8.2 & 3.5$\pm$0.9 & 3.5$\pm$1.0& 3.5$\pm$1.1& 3.6$\pm$1.0& 2.9$\pm$1.1\\
    Best DUC system (LQ) & -- & 4.0$\pm$0.8 & 4.2$\pm$0.7& 3.8$\pm$0.7& 3.6$\pm$0.9& 3.4$\pm$0.9\\
    Our System & {\bf 26.4}$\pm$10.3 & 3.0$\pm$0.9& 4.0$\pm$1.1& 3.6$\pm$1.0& 3.4$\pm$0.9& 2.8$\pm$1.0\\
	
	\hline
	\end{tabular}
    }
    \caption{\fontsize{10}{12}\selectfont Human evaluation on our multi-scorer based system, \newcite{iiit06} (Best DUC system (ROUGE)), and \newcite{lcc06} (Best DUC system (LQ)). Our system can synthesize more relevant content according to Pyramid ($\times 100$). We also examine linguistic quality (LQ) in Grammaticality (Gra), Non-redundancy (Non-Red), Referential clarity (Ref), Focus (Foc), and Structure and Coherence (Coh) like~\newcite{duc}, each rated from 1 (very poor) to 5 (very good). Our system has better non-redundancy than~\newcite{iiit06} and is comparable to~\newcite{iiit06} and~\newcite{lcc06} in other metrics except grammaticality.}
    \label{tab:resultHuman}
\end{table*}

\begin{figure}[ht]
\centering
    {\footnotesize
    \setlength{\baselineskip}{0pt}
    \begin{tabular}{|p{72mm}|}
    \hline
    Topic $D0626H$: How were the bombings of the US embassies in Kenya and Tanzania conducted?  What terrorist groups and individuals were responsible? How and where were the attacks planned?\\
    \hline
    \end{tabular}
    
    \begin{tabular}{|p{72mm}|}
    \hline
    \textcolor{Gray}{WASHINGTON, August 13 (Xinhua) --} President Bill Clinton \textcolor{Gray}{Thursday} condemned terrorist bomb attacks at U.S. embassies in Kenya and Tanzania and vowed to find the bombers and bring them to justice.
   Clinton met with his top aides \textcolor{Gray}{Wednesday in the White House} to assess the situation following the twin bombings at U.S. embassies in Kenya and Tanzania\textcolor{Gray}{, which have killed more than 250 people and injured over 5,000, most of them Kenyans and Tanzanians}.
    \textcolor{Gray}{Local sources said} the plan to bomb U.S. embassies in Kenya and Tanzania took three months to complete and bombers \textcolor{Gray}{destined for Kenya} were dispatched \textcolor{Gray}{through Somali and Rwanda}.
	FBI Director Louis Freeh, Attorney General Janet Reno and other senior U.S. government officials \textcolor{Gray}{will} hold a news conference at 1 p.m. \textcolor{Gray}{EDT (1700GMT)} at FBI headquarters in Washington \textcolor{Gray}{``}to announce developments in the investigation of the bombings of the U.S. embassies in Kenya and Tanzania\textcolor{Gray}{," the FBI said in a statement}. ...
	\\
	\hline
	\end{tabular}
    }
	\vspace{-1mm}
    \caption{\fontsize{10}{12}\selectfont Part of the summary generated by the multi-scorer based summarizer for topic $D0626H$ (DUC 2006). Grayed out words are removed. Query-irrelevant phrases, such as temporal information or source of the news, have been removed.}
    \label{fig:sampleSummary}

\end{figure}

The Pyramid~\cite{nenkova-passonneau:2004:HLTNAACL} evaluation was developed to manually assess how many relevant facts or Summarization Content Units (SCUs) are captured by system summaries. We ask a professional annotator (who is not one of the authors,  is highly experienced in annotating for various NLP tasks, and is fluent in English) to carry out a Pyramid evaluation on 10 randomly selected topics from the DUC 2006 task with gold-standard SCU annotation in abstracts. The Pyramid score (see Table~\ref{tab:resultHuman}) is re-calculated for the system with best ROUGE scores in DUC 2006~\cite{iiit06} along with our system by the same annotator to make a meaningful comparison.

We further evaluate the linguistic quality (LQ) of the summaries for the same 10 topics in accordance with the measurement in~\newcite{duc}. Four native speakers who are undergraduate students in computer science (none are authors) performed the task,
We compare our system based on {\sc Head}-driven beam search with {\sc Multi}-scorer to the best systems in DUC 2006 achieving top ROUGE scores~\cite{iiit06} (Best DUC system (ROUGE)) and top linguistic quality scores~\cite{lcc06} (Best DUC system (LQ))\footnote{\newcite{lcc06} obtain the best scores in three linguistic quality metrics (i.e.~grammaticality, focus, structure and coherence), and overall responsiveness on DUC 2006.}.
The average score and standard deviation for each metric is displayed in Table~\ref{tab:resultHuman}. Our system achieves a higher Pyramid score, an indication that it captures more of the salient facts. We also attain better non-redundancy than~\newcite{iiit06}, meaning that human raters perceive less replicative content in our summaries. Scores for other metrics are comparable to~\newcite{iiit06} and~\newcite{lcc06}, which either uses minimal non-learning-based compression rules or is a pure extractive system. However, our compression system sometimes generates less grammatical sentences, and those are mostly due to parsing errors. For example, parsing a clause starting with a past tense verb as an adverbial clausal modifier can lead to an ill-formed compression. Those issues can be addressed by analyzing $k$-best parse trees and we leave it in the future work.
A sample summary from our multi-scorer based system is in Figure~\ref{fig:sampleSummary}.

\paragraph{Sentence Compression Evaluation.} 
%

\begin{table*}[ht]
\centering
    {\small
    \setlength{\baselineskip}{0pt}
    \begin{tabular}{|l|c|c|c|c|c|}
    \hline
    \textbf{System}&\textbf{C Rate}&\textbf{Uni-Prec}&\textbf{Uni-Rec}&\textbf{Uni-F1}&\textbf{Rel-F1}\\ \hline
    HedgeTrimmer & 57.64\%& 0.72 & 0.65 & 0.64 & 0.50\\
    McDonald (2006)& 70.95\% & 0.77 & 0.78 & {\it 0.77} & 0.55\\
	\newcite{Martins:2009:SJM:1611638.1611639}& 71.35\% & 0.77 & 0.78 & {\it 0.77} & 0.56\\
	
	\hline
	Rule-based &87.65\% & 0.74 & 0.91 & 0.80 & 0.63\\
	Sequence &70.79\% & 0.77 & 0.80 & {\it 0.76} & 0.58\\
	Tree ({\sc Basic}) & 69.65\% & 0.77 & 0.79 & 0.75 & 0.56\\
	Tree ({\sc Context}) & 67.01\% & {\bf 0.79} & 0.78 & {\it 0.76} & 0.57\\
	Tree ({\sc Head}) & 68.06\% & {\bf 0.79} & 0.80 & {\it 0.77} & {\bf 0.59}\\
	\hline
	\end{tabular}
    }
    \caption{\fontsize{10}{12}\selectfont Sentence compression comparison. The true c rate is 69.06\% for the test set. Tree-based approaches all use single-scorer. Our context-aware and head-driven tree-based approaches outperform all the other systems significantly ($p<0.01$) in precision ({\bf Uni-Prec}) without sacrificing the recalls (i.e.~there is no statistically significant difference between our models and McDonald (2006) / M \& S (2009) with $p>0.05$). {\it Italicized} numbers for unigram F1 ({\bf Uni-F1}) are statistically indistinguishable ($p>0.05$). Our head-driven tree-based approach also produces significantly better grammatical relations F1 scores ({\bf Rel-F1}) than all the other systems except the rule-based method ($p<0.01$).}
    \label{tab:resultComp}
\end{table*}

We also evaluate sentence compression separately on~\cite{Clarke:2008:GIS:1622655.1622667}, adopting the same partitions as~\cite{Martins:2009:SJM:1611638.1611639}, i.e.~$1,188$ sentences for training and $441$ for testing. 
Our compression models are compared with Hedge Trimmer~\cite{Dorr2003}, a discriminative model
proposed by~\newcite{McDonald06} and a dependency-tree based 
compressor~\cite{Martins:2009:SJM:1611638.1611639}\footnote{Thanks to Andr{\'e} F.T. Martins for system outputs.}. 
We adopt the metrics in~\newcite{Martins:2009:SJM:1611638.1611639} to measure the unigram-level
macro precision, recall, and F1-measure with respect to human annotated compression. In addition, we also compute the F1 scores of grammatical relations which are annotated by RASP~\cite{RASP} according to~\newcite{Clarke:2008:GIS:1622655.1622667}.

In Table~\ref{tab:resultComp}, our context-aware and head-driven tree-based compression systems show statistically significantly ($p<0.01$) higher precisions ({\bf Uni-Prec}) than all the other systems, without decreasing the recalls ({\bf Uni-Rec}) significantly ($p>0.05$) based on a paired {\it t}-test. Unigram F1 scores ({\bf Uni-F1}) in italics indicate that the corresponding systems are not statistically distinguishable ($p>0.05$). For grammatical relation evaluation, our head-driven tree-based system obtains statistically significantly ($p<0.01$) better F1 score ({\bf Rel-F1} than all the other systems except the rule-based system).



\section{Conclusion}
We have presented a framework for query-focused multi-document summarization based on sentence compression. We propose three types of compression approaches. Our tree-based compression method can easily incorporate measures of query relevance, content importance, redundancy and language quality into the compression process. By testing on a standard dataset using the automatic metric ROUGE, our models show substantial improvement over pure extraction-based methods and state-of-the-art systems. Our best system also yields better results for human evaluation based on Pyramid and achieves comparable linguistic quality scores.

\section*{Acknowledgments}
This work was supported in part by National Science Foundation Grant IIS-0968450 and a gift from Boeing. We thank Ding-Jung Han, Young-Suk Lee, Xiaoqiang Luo, Sameer Maskey, Myle Ott, Salim Roukos, Yiye Ruan, Ming Tan, Todd Ward, Bowen Zhou, and the ACL reviewers for valuable suggestions and advice on various aspects of this work.

{\small
\bibliographystyle{acl2013}

}
\end{document}